\documentclass[10pt,twocolumn,letterpaper]{article}

\usepackage[pagenumbers]{cvpr} 

\usepackage{algorithm}
\usepackage{algorithmic}
\usepackage{epsfig}
\usepackage{amsmath}
\usepackage{dsfont}
\usepackage{amssymb}
\usepackage{mathrsfs}
\usepackage{booktabs}
\usepackage{xcolor}
\usepackage{pifont}
\newcommand{\cmark}{\ding{51}}%
\newcommand{\xmark}{\ding{55}}%
\usepackage{multirow}
\usepackage[normalem]{ulem}
\graphicspath{{figures/}}
\usepackage{relsize}
\usepackage{amsthm,enumitem}
\usepackage{footnote}
\usepackage{blindtext}
\usepackage{graphicx}

\usepackage[pagebackref,breaklinks,colorlinks]{hyperref}

\usepackage[capitalize]{cleveref}
\crefname{section}{Sec.}{Secs.}
\Crefname{section}{Section}{Sections}
\Crefname{table}{Table}{Tables}
\crefname{table}{Tab.}{Tabs.}

\def\cvprPaperID{5958} 
\def\confName{CVPR}
\def\confYear{2022}

\makesavenoteenv{tabular}
\makesavenoteenv{table}

\begin{document}

\title{SA-VQA: Structured Alignment of Visual and Semantic Representations for Visual Question Answering}

\author{\textbf{Peixi Xiong$^{1}$}\thanks{Work done during an internship at Microsoft Research.}\textbf{, Quanzeng You$^{2}$, Pei Yu$^{2}$, Zicheng Liu$^{2}$, Ying Wu$^{1}$}\\
${}^{1}$Northwestern University, ${}^{2}$Microsoft Research
}
\maketitle

\begin{abstract}
Visual Question Answering (VQA) attracts much attention from both industry and academia. 
As a multi-modality task, it is challenging since it requires not only visual and textual understanding, but also the ability to align cross-modality representations.
Previous approaches extensively employ entity-level alignments, such as the correlations between the visual regions and their semantic labels, or the interactions across question words and object features.
These attempts aim to improve the cross-modality representations, while ignoring their internal relations.
Instead, we propose to apply structured alignments, which work with graph representation of visual and textual content, aiming to capture the deep connections between the visual and textual modalities. 
Nevertheless, it is nontrivial to represent and integrate graphs for structured alignments. 
In this work, we attempt to solve this issue by first converting different modality entities into sequential nodes and the adjacency graph, then incorporating them for structured alignments. 
As demonstrated in our experimental results, such a structured alignment improves reasoning performance. 
In addition, our model also exhibits better interpretability for each generated answer. 
The proposed model, without any pretraining, outperforms the state-of-the-art methods on GQA dataset, and beats the non-pretrained state-of-the-art methods on VQA-v2 dataset.
\end{abstract}

\section{Introduction}
Recently, Visual Question Answering (VQA) has received much attention in both industry and academia.
This can be partially attributed to its broad applications (\eg, early educational systems). 

Even deep learning models achieve increasingly better performance across many computer vision (CV) and natural language processing (NLP) tasks, their performance may still be inadequate for applications with diversified testing scenarios.
This also happens in VQA. 
It is already challenging for models to understand the concepts in visual or textual content individually. 
Moreover, this task further requires cross-modality understanding. 
To achieve competitive performance in VQA, machine learning models need to represent and align different entities in both visual and textual content.
Entity-level alignments are insufficient for VQA, since the task usually requires the model to detect the deep connections between these two modalities. 
We argue structured alignments are necessary for deep cross-modality understanding. 
However, this requires a graph representation of visual and textual content. 


\begin{figure}[t]
	\centering
	\includegraphics[width=0.9\columnwidth]{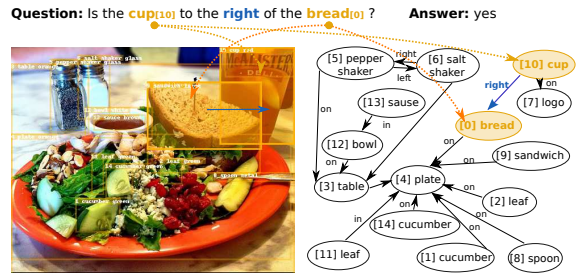}
		\caption{Example of an image and its scene graph. Given the question, human can easily match the keywords of questions to image regions and scene graph entities, from which we can infer the answer.}
		\label{fig:figure1}
\end{figure}



Specifically, in the CV domain, such a graph is defined as the scene graph, a structured graphical representation of the image.
Each scene graph encodes high-level visual concepts, including objects, attributes, and relations, as nodes and edges.
This type of structured representation has led to many state-of-the-art models in image captioning~\cite{yang2019auto}, image retrieval~\cite{johnson2015image}, relationship modeling~\cite{woo2018linknet}, and image generation~\cite{johnson2018image}.
In addition, it promotes the publishing of several datasets containing well-annotated scene graphs, such as GQA~\cite{DBLP:conf/cvpr/HudsonM19} and Visual Genome~\cite{krishnavisualgenome}, to facilitate research on the use of scene graphs for visual and textual content understanding.

\figurename~\ref{fig:figure1} shows the pair of an image and its corresponding scene graph. 
Given the constructed graph, it is straightforward for a human to infer that the answer to the presented question is ``yes''. 
However, to allow VQA methods to leverage the information from scene graphs for reasoning, a challenging problem must be addressed:
``\textit{Is there any way to embed structured graph information into the learning and reasoning process of VQA models?}''
More specifically, the problem can be formulated as how to align features from the structured scene graphs, visual content, and questions, and how to reason the answer from them.
Taking \figurename~\ref{fig:figure1} as an example, when the objects and relations in the scene graph are well aligned with the question, it is straightforward for the model to infer the answer.

In previous studies~\cite{teney2017graph,DBLP:conf/nips/Norcliffe-Brown18}, fully connected scene graphs or conditional relational graphs were constructed from either visual concepts or questions. 
Such structured representation is further processed by Graph Neural Networks (GNNs), which usually apply message passing for graph modeling. 
However, GNNs are inefficient in terms of updating the node hidden states, and the edges are also tricky to model~\cite{zhou2018graph}.
In addition, the results lack interpretability when only features are embedded and passed to neighbors.
Nevertheless, these do not imply that the GNNs will be replaced in all graph-based tasks. 
Instead, many ongoing studies on GNNs are conducted to overcome their limitations and achieve better performance. 
One direction is to combine GNNs and Transformers to model both local and global interactions~\cite{lin2021mesh}.

In this work, we leverage Transformer to tackle this problem in a unified framework.
The attention module in Transformer allows us to disclose correlations between input tokens, which is critical for modality alignment and understanding. 
Thus, it has been widely employed for cross vision and language tasks.  
The conventional attention modules learned correlations between different input tokens fully from data. 
These modules can attain meaningful outcomes when the task requires less reasoning and large-scale training data is presented. 
However, the learning can be extremely slow for tasks that demand intensive reasoning, and building the large-scale training dataset can also be challenging. 
Instead, we propose to use graph-structured information to guide learning. 
With the help of structured information, it is possible to learn the deep alignments between visual and textual representation, even when presented with limited training data. 
Extensive experiments demonstrate the superiority of embedding graph information for visual question answering. Our main contributions are as follows:
\begin{itemize}
	\item We propose a novel model that learns with Structured Alignment of visual and semantic representations for Visual Question Answering (SA-VQA), which is capable of learning correlations from the visual and textual content.
	\item We experiment on both datasets GQA~\cite{DBLP:conf/cvpr/HudsonM19} and VQA-v2~\cite{DBLP:conf/cvpr/HudsonM19}. SA-VQA outperforms other non-pretrained state-of-the-art methods.
	\item We conduct ablation studies by examining different modules. Our results demonstrate the effectiveness of each module. Extra experimental results further reveal the potential capacity of SA-VQA when presented with a better semantic graph.
\end{itemize}

\begin{figure*}[tb!] 
\centerline{\includegraphics[width=0.9\textwidth]{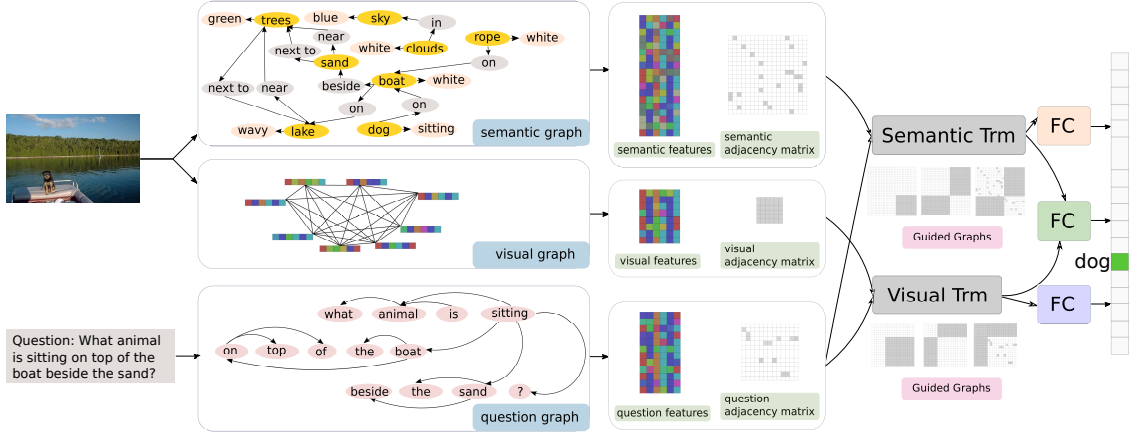}}
\caption{Overview of our approach. 
Objects' labels, attributes, and relations are detected from the image, represented by the yellow, orange, and gray nodes, respectively. They form the semantic graph based on their connectivity.
The visual graph and question graph are built in a similar way.
These three graphs are then converted into input sequences together with their corresponding adjacency matrices.
The three sequences are inputs to two Transformers (Trms), where the adjacency matrices act as structured information to the Transformers.
The outputs of the two Transformers are then used to predict the final answer.}

\label{fig-overview}
\end{figure*}

\section{Related works}
In this section, we discuss the related studies on using graphs for VQA and recent developments of Transformers. 
\subsection{Graph structures in VQA reasoning}
The reasoning for VQA requires an understanding of both visual content and natural language concepts. 
Many works~\cite{DBLP:journals/pr/YuZWZHT20, DBLP:conf/nips/Yi0G0KT18} have focused on this topic from different perspectives.
Compositional models for reasoning~\cite{DBLP:journals/corr/AndreasRDK15} break down natural language queries and visual understanding into several steps; alternatively, in many cases, they involve much extra manual effort to define the task's substeps. 
Dynamic memory networks~\cite{DBLP:conf/icml/XiongMS16} and attention mechanisms~\cite{DBLP:conf/cvpr/NamHK17} are two leading solutions for addressing tasks requiring complex logical reasoning because they involve interactions among multiple data components.

All these works have improved VQA in performing reasoning. Researchers have found that graphs also play an important role in reasoning for VQA since they preserve concepts' structural information. 
However, the challenge is that images lack a language structure and grammatical rules~\cite{DBLP:journals/corr/WuTWSDH16}. 
Motivated by this and the demand for better image understanding, research on the construction of graphs to represent image information has been undergoing rapid development~\cite{DBLP:conf/cvpr/ZhangECCE17}. 
Many research efforts~\cite{teney2017graph, DBLP:journals/corr/abs-2101-05479, DBLP:journals/corr/abs-1907-12133, DBLP:conf/gc/LeeKOJ19} focus on the important role of scene graphs. 
However, constructed scene graphs may also contain errors that require additional effort to handle.


\subsection{Transformers}
The Transformer architecture~\cite{NIPS2017_3f5ee243} was initially
proposed for language modeling. 
It has (1) an attention module using a scaled dot product and (2) a fast training time due to its parallel architecture. 
The model not only can effectively address machine translation tasks, but also can be used for vision tasks, such as object detection~\cite{DBLP:conf/eccv/CarionMSUKZ20}, image enhancement~\cite{DBLP:journals/corr/abs-2012-00364}, image generation~\cite{DBLP:conf/icml/ParmarVUKSKT18}, segmentation~\cite{DBLP:journals/corr/abs-2012-00759}, and learning useful image representations~\cite{DBLP:conf/icml/ChenRC0JLS20}. %
It achieves better multi-modality alignment when used in visual language tasks~\cite{tan2019lxmert, DBLP:journals/corr/abs-2010-12831,DBLP:conf/iclr/SuZCLLWD20}. 
However, in contrast to our work, these approaches usually rely on pre-training with a large amount of data~\cite{NEURIPS2019_98d8a23f} and computational resources to achieve better performance. 

\section{Approach overview}
\label{section-overview}
We now introduce the details of the proposed model SA-VQA.
\figurename~\ref{fig-overview} illustrates the architecture of the proposed model, which consists of two parallel streams: visual and semantic.
First, objects are detected, along with the object labels, object attributes, and their relations.
Next, three different graphs, \ie, semantic graph, visual graph, and question graph, are constructed.
\figurename~\ref{fig-overview} shows an example of the three graphs. 
We believe semantic and visual graphs are complementary to each other. The semantic graph expedites the integration of structured features, and the visual graph offers supplementary information, which may be misplaced in the semantic graph.
The question graph further allows the inference to pay extra attention to the hints embedded in the question.
They will be inputs to the visual and semantic streams, which employ Transformers to highlight the correlations between the question and the corresponding image.
Finally, the outputs of the two Transformers are used to predict the answer with a multi-layer perceptron (MLP) classifier.

Section~\ref{section-g} describes how to extract the semantic and visual graph from the input image and the question graph from the input question. 
Section~\ref{section-t} explains in detail the usage of the embedded graphs and the proposed mechanism to work with different graphs to achieve structured alignment. 
\section{Graph construction}
\label{section-g}
The structured representation is described by the ER-model~\cite{chen1976entity} with the entity set $\mathcal{V}$, the set of relations $\mathcal{P}$ between pairs of entities, and the attribute set $\mathcal{A}$. 
Then, an image can be represented by a set of triplets $(e_s, e_p, e_o)$. The subject $e_s$ is an entity in $\mathcal{V}$: $e_p$ is a predicate in $\mathcal{P}$, and object $e_o$ is either an entity in $\mathcal{V}$ or an attribute value $a$ in $\mathcal{A}$. 
Our modality graph is defined as $\mathcal{G} = (\mathcal{E}, \mathcal{R})$, where $e_s$ and $e_o$ are regarded as $\mathcal{E}$, and the relations $e_p$ from $e_s$ to $e_o$ is regarded as $\mathcal{R}$.

From the given image, we construct two separate graphs: \textit{semantic} and \textit{visual} graph. 
\figurename~\ref{fig-overview} shows an example of the two graphs as well as a question graph.
The details are discussed in this section.
\subsection{Semantic graph}
\begin{figure}[!htbp]
	\centering
	\includegraphics[width=0.8\columnwidth]{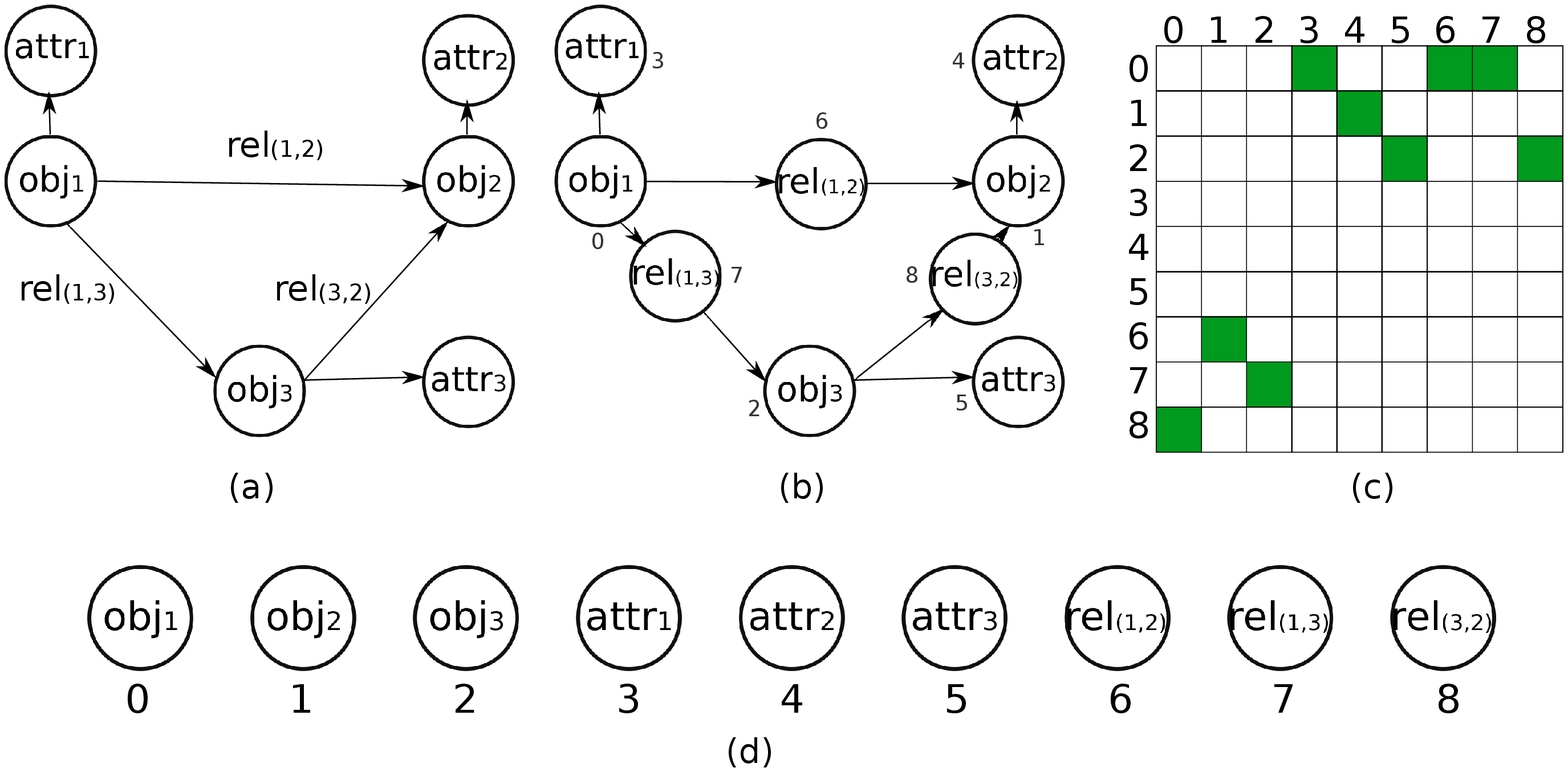}
	\caption{Graph conversion. 
		(a) The graph extracted from the image. $\text{obj}_i$ represents the $i$-th detected object, $\text{attr}_i$ is the set of its attributes (\eg, color and shape), and $\text{rel}_{(i,j)}$ is the relation between $\text{obj}_i$ and $\text{obj}_j$. 
		(b) $\text{rel}_{(i,j)}$ is also regarded as a node in the graph, and each node in the graph is assigned a numeric index. 
		Then, the graph can be represented as (c) the adjacency matrix that indicates the connectivity of the nodes in (d), where the green cell and the blank cell represent  values of 1 and 0 respectively.}
	\label{fig:graph_convert}
\end{figure}
For a given image, we first recognize the objects, attributes, and the relations between object pairs. 
Then, we form a \textit{semantic} graph $\mathcal{G}_s = (\mathcal{E}_s, \mathcal{R}_s)$ by regarding $e_s$ and $e_o$ as the entity $\mathcal{E}_s$, and the predicate $e_p$ as the edge $\mathcal{R}_s$. 
We also consider the relations as nodes in our feature sequence to better use the constructed graph as the input to the next stage and the subjects, objects, and attributes. 
In this way, we convert the graph into a sequence of nodes $\{n_i\}_{i=1}^{N}$ and a corresponding adjacency matrix $A_s$.
This is illustrated in detail in \figurename~\ref{fig:graph_convert}.

The sequence of the node features  $\{s_1, s_2, ..., s_{N}\}$ is computed from the node sequence $\{n_i\}_{i=1}^{N}$ by GloVe embedding~\cite{Pennington14glove:global} processed by a Multi-layer perceptron (MLP). 

\subsubsection{SuperNode selection (SNS) over bounding box label candidates} 

Scene graph generation has been a very challenging research task.
This task itself is beyond the scope of this work.
We employ an off-the-shelf object and attribute model~\cite{anderson2018bottom} to discover the nodes and edges that are utilized to construct the semantic graph. 
More details will be discussed in our experiment settings.

However, inaccurate objects and attributes will result in poor semantic graphs, which may mislead the learning of our model. 
To reduce the impact of poor semantic graphs, we use all the top-K prediction results for each candidate bounding box, regarded as a \textbf{SuperNode}.  
	SuperNode covers redundant object nodes, and the recall of objects is much higher as well. 
	This is necessary, as the goal of VQA model is to infer the answers from relevant features. 
	If a relevant object node is absent from the semantic graph, learning from an incomplete graph would be more challenging. 
	Indeed, even human beings may be incapable of recovering the answer when solely presented with an incomplete semantic graph.
	
	
	However, each SuperNode also introduces noisy and irrelevant information. 
	Extra mechanisms are demanded to select the most relevant information related to the given question.
	A tentative solution is to learn to assign different weights for each object in the SuperNode.
	Then, each semantic feature of SuperNode ($f_{\text{SNS}_i}$) can be represented as the weighted semantic features $f_n$ of each object within the set:
	\begin{equation}
		\begin{aligned}
			f_{\text{SNS}_i} & = \sum_{n_j \in \text{SNS}_i} w_{ij}f_{n_j}\\
			w_{ij} & = \frac{\exp(f_{n_j}^Tv_i)}{\sum_{n_j' \in \text{SNS}_i} \exp(f_{n_j'}^Tv_i)}
		\end{aligned},
	\end{equation}
	where $w_{ij}$ is the normalized learnable weight of node $n_j$ in the set of nodes $\text{SNS}_i$ and $v_i$ is the visual feature of the $i$-th detected bounding box.
	Accordingly, we can follow the same graph conversion steps discussed in the previous subsection to convert graphs with SuperNodes.
	
	\noindent\textbf{SuperNode selection (SNS)} \quad We employ SuperNode selection (SNS) to learn the weights $w_{ij}$ for each node $n_j$ in a SuperNode $\text{SNS}_i$. 
	The input for SNS consists of the top-K semantic features $f_{n_j}$ of each object label within each $\text{SNS}_i$.
	Our goal is to assign a larger weight for object label that matches ground truth and smaller weights for others.
	
	To achieve this, we design two parallel tasks. 
	The first is to bring the top-K semantic features ($\{f_{n_{j}}\}_{j=1}^{K}$) closer to the visual feature $v_i$ of the $i$-th object region than the other semantic features $f_{n'}$ that are not in the top-K set. 
	
	This task falls within the scope of Multiple Instance Learning (MIL) settings. 
	In particular, we employ MIL-NCE~\cite{Miech_2020_CVPR} to perform this task. 
	The positive candidates are the top-K features, while the negative candidates are semantic features of those object labels $\{n'\}$ that are randomly sampled from the object vocabulary.
	
	The second task is designed for nodes within the set of SuperNode $\text{SNS}_i$.
	The objective is to bring the ground-truth feature $f_{n^i_{\text{GT}}}$ closer to the visual object region feature than the other features $f_{n_j}$ ($n_j \ne n^i_{\text{GT}}$) within the top-K set. 
	For this task, we employ the contrastive loss $\mathcal{L}_\text{CL}$ as the loss function, where the positive sample is the ground truth, and others within the top-K set are negative samples.
	
	In addition, since distributed representations of words usually demonstrate analogies~\cite{Pennington14glove:global}, we compute the L2 distance between the weighted semantic features $f_{\text{SNS}_i}$ to the ground-truth semantic feature $f_{n^i_{\text{GT}}}$ to further assist the learning of the weights in each SuperNode.
	In summary, our objective in SNS is to obtain the optimal visual feature $v_i$ and semantic feature $f_n$ by minimizing the following loss function:
	\begin{equation}
		\begin{aligned}
			\mathcal{L}_{\text{SNS}} & = \mathcal{L}_{\text{MIL}}^{\textbf{$\mathds{K}$}}
			+ \mathcal{L}_\text{CL}
			+ \sum_i D\left(f_{n^i_{\text{GT}}}, f_{\text{SNS}_i}\right) \\
			\mathcal{L}_{\text{MIL}}^{\mathds{K}} &= \\
			- \sum_{i=1}^{N} &\log \frac{\sum_{{n_j}\in \mathcal{P}_i} \exp({f_{n_j}^Tv_i})}
			{\sum\limits_{{n_j}\in \mathcal{P}_i} \exp(f_{n_j}^Tv_i) + 
				\sum\limits_{{n_j'} \in \mathcal{N}_i} \exp(f_{n_j'}^Tv_i)}
		\end{aligned}
	\end{equation}
	Here, $i$ indexes over all bounding boxes, and the total number of boxes is $N$. 
	$\mathcal{P}_i$ represents the set of the index of the top-K detected object labels for the $i$-th bounding box, and $v_i$ is the visual feature. Similarly, $\mathcal{N}_i$ represents the set of randomly sampled negative labels for the $i$-th bounding box.
	$D(\cdot, \cdot)$ calculates the L2 distance between two feature vectors.
	%
	%
	%
	%
	%
	
	\subsection{Visual graph}
	To embed the visual features of images, we construct the visual graph $\mathcal{G}_v = (\mathcal{E}_v, \mathcal{R}_v)$, where $\mathcal{E}_v$ denotes the detected regions by Faster-RCNN~\cite{anderson2018bottom}, represented by their regional features from ROIAlign~\cite{he2017mask}. We choose to keep all the pair-wise connections between all nodes, which makes $\mathcal{G}_v$ fully connected. 
	This graph, which includes redundant connections, intends to compensate for incorrect and missing links in the semantic graph $\mathcal{G}_s$.
	
	The conversion of the visual graph follows a similar pipeline with the previously discussed semantic graph (see \figurename~\ref{fig:graph_convert} for an example).
	Differently, visual graph conversion is simpler, which includes $M$ detected objections with their feature sequence $\{v_1, v_2, ..., v_M\}$ and the fully connected adjacency matrix $A_v$.
	
	\subsection{Question graph}
	Tree structures, such as \textit{syntax tree} and \textit{parsing tree}, have been widely studied and exploited across different NLP tasks. 
	In this work, we employ the dependency parsing tree~\cite{kubler2009dependency} to generate the graph $\mathcal{G}_q = (\mathcal{E}_q, \mathcal{R}_q)$ for each question $q_i$.
	
	In particular, we represent each node by GloVe embedding~\cite{Pennington14glove:global} followed by an MLP. 
	The corresponding binary adjacency matrix $A_q$ represents the edges in the dependency parsing tree, where $A_q(i,j) = 1$ indicates two nodes are connected in the dependency tree. 

\section{Structured representation integration}
\label{section-t}

As discussed in Section~\ref{section-overview}, we adopt Transformer for modeling. 
This section will exhibit how to integrate the above-constructed graphs into our model learning.

\subsection{Attention learning for structured alignment}

\begin{figure}[!htbp]
	\centering
	\includegraphics[width=0.9\columnwidth]{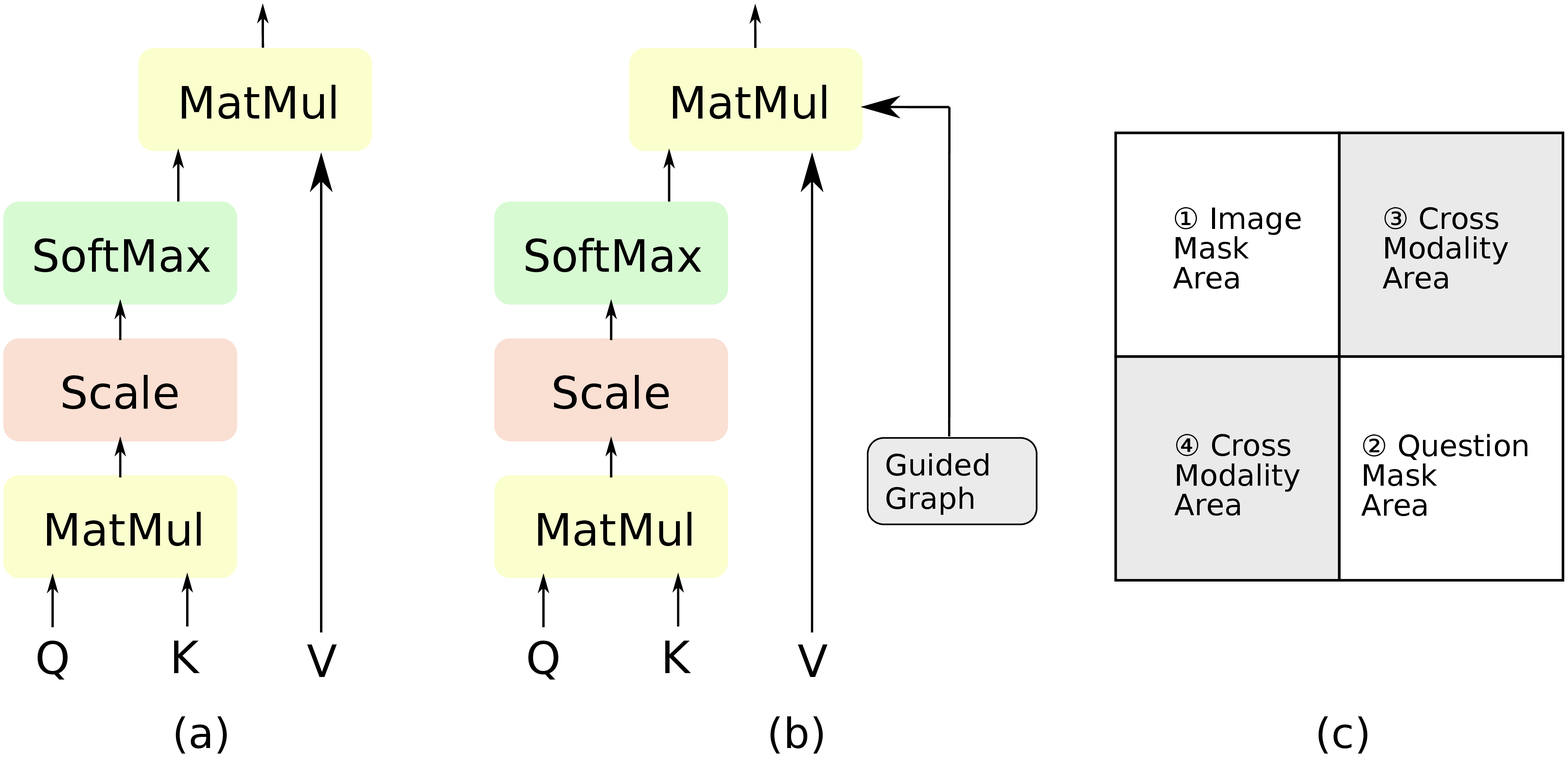}
		\caption{Different attention mechanisms in Transformer. 
			(a) The scaled dot-product attention {initially proposed in} \cite{NIPS2017_3f5ee243} . 
			(b) The guided graph $G$ is applied to the attention.
			(c) Adjacency matrix of graph $G$ is divided into four parts. The diagonal parts are used for attention on the image (Region 1) and the question (Region 2) separately, while the opposite diagonal parts focus on cross-modality attention (Regions 3 and 4).}
		\label{fig:attention}
\end{figure}
The Transformer architecture~\cite{NIPS2017_3f5ee243} consists of the encoder and the decoder, both of which apply multi-head attention module for alignment learning.
In both the encoder and decoder, scaled dot-product attention (\figurename~\ref{fig:attention}) is utilized to compute the output from the given query and key-value pairs. 
The attention mechanism can discover the connections embedded in the input tokens. 
It demonstrates its success across a wide range of NLP and multi-modality alignment tasks. 
However, it usually requires a large amount of training data for the model to converge and achieve decent performance.

Instead, our goal is to reduce the dependency on large-scale training data. 
We attempt to assist the learning of attention modules by incorporating structured graph features.
We believe that the external graph features, even incomplete and inaccurate, can still scale down the searching space of the attention module. 
Therefore, it is capable of learning feature relations even presented with limited training data.

To integrate the structured graph feature, we propose a simple yet effective design as shown in \figurename~\ref{fig:attention}~(b).
More specifically, the attention computation is forced to consider the graph $G$ as an extra constraint:

{
	\begin{equation}
		\mathsmaller{\text{Att}_{g}(Q, K, V, G) = h\left(\text{softmax}(\frac{QK^T}{\sqrt{ d_k }})\odot G\right)~V},
	\end{equation}
}

\noindent where $d_k$ represents the dimensionality of the input, {$\odot$ denotes element-wise multiplication}, {$h(X)$ normalizes the rows of $X$ to sum to one (\ie, $h(X)_{i, j}=\frac{X_{i,j}}{\sum_{j}X_{i,j}}$}), and G is composed from the binary adjacency matrix, \ie, $A_s$, $A_v$ and $A_q$.  The following sections show more details.

\subsection{Visual Transformer ($T_v$)}
The input feature sequence to $T_v$ is $\{f^v_i\}_{i=0}^M$, which is the concatenation of the two feature sequences from $\mathcal{G}_v$ and $\mathcal{G}_q$. 
Similar to~\cite{NIPS2017_3f5ee243}, positional encoding is {employed} to combine the relative or absolute position information with the learned positional embeddings.

The graph constraints for $T_v$ are of three types: the question adjacency graph $A_{q}$, cross-modality mask $g_{\text{cm}}$, and visual adjacency graph $A_v$. 
Assuming the image feature sequence is length $V$ and the question feature sequence is length $Q$, we have $M = V + Q$. 
\figurename~\ref{fig:attention}~(c) shows the composition of the constraint graph $G$. 
\textit{Region 1} is the visual graph $A_v\in \mathbb{R}^{V\times V}$.
\textit{Region 2} represents $A_q\in \mathbb{R}^{Q\times Q}$. \textit{Regions 3} and \textit{4} mainly focus on the attention across the image and the question.
Firstly, we only set \textit{Region 2} of $G$ to $A_q$ to guide the learning of better question embedding. 
Next, we reset $G$ and \textit{Region 3} and \textit{Region 4} are filled with $1$, which forces the encoder to learn the cross attention between features from the images and questions.
Lastly, we reset $G$ to include \textit{Region 1} ($A_v$), \textit{Region 2} ($A_q$), \textit{Region 3} and \textit{Region 4} as the constraint. 
This makes the encoder focus on the existing connectivity in the visual graph and question graph.
These settings are similar to~\cite{NEURIPS2019_c74d97b0}, which addresses the question features first, transforming them into the same feature space as other modalities.
\subsection{Semantic Transformer ($T_s$)}
The semantic branch {(see \figurename~\ref{fig-overview})} is similar to the visual branch. 
However, instead of using visual features, the input representation is from semantic graph $\mathcal{G}_s$.
The graph constraint $G$ is constructed similarly to Visual Transformer $T_v$ in the previous section, consisting of constraints for question self-attention, cross-modality attention, and adjacency connectivity.

\subsection{Output representation}
As shown in the rightmost part of \figurename~\ref{fig-overview}, our model has two streams of output from $T_s$ and $T_v$. 
Unlike sequence-to-sequence learning tasks, our model regards VQA task as a classification problem, and uses Fully-Connected Layers (FCs) to process Transformer outputs.
We compute cross-entropy loss from the visual and semantic streams individually.
We also use both early fusion and late fusion strategies to fuse two types of features for classification.
Overall, our training loss is defined as follows:
\begin{equation}
	\mathsmaller{L(\mathcal{G}_s, \mathcal{G}_v, \mathcal{G}_q, a) = {L_{\text{CE}}}(f_v, a) + {L_{\text{CE}}}(f_s, a) + {L_{\text{CE}}}(f_f, a)}
	\label{eq:loss}
\end{equation}
\noindent{where $f_v$, $f_s$ and $f_f$ represent the logits for visual Transformer, semantic Transformer and their concatenation, respectively, and $a$ is the answer to the question}.

\section{Experimental setup and results}
\label{section-exp}
\subsection{Datasets}
\subsubsection{VQA-v2}
The VQA-v2~\cite{DBLP:journals/ijcv/AgrawalLAMZPB17} dataset includes 204,721 images from the COCO dataset and 1,105,904 open-ended questions related to images. 
Each question in the dataset is associated with ten different answers.
The evaluation metric, which is robust to inter-human variability, is $acc(ans) = min\{\frac{\#~ans~is~chosen}{3}, 1\}$ (\ie, an answer is deemed 100\% accurate if at least 3 annotated answers exactly match that answer).

\subsubsection{GQA}
The GQA~\cite{DBLP:conf/cvpr/HudsonM19} dataset is a large-scale dataset that addresses previous VQA datasets' key shortcomings, \ie, the visual questions are believed to be biased and relatively simple.
Differently, the GQA dataset attempts to collect more complicated and compositional questions, which require more reasoning over the visual concepts. 
It consists of 22M novel and diverse questions and 113k images. 
Its questions display a richer vocabulary and more various linguistic and grammatical structures than other VQA datasets.
Moreover, it contains clean, unified, rich, and unambiguous scene graphs (including bounding box information, attributes, and relations) for 85k+ images (training and validation splits only), which are normalized and extended from the Visual Genome~\cite{krishnavisualgenome} dataset. 
We evaluated our model on the testing split of the GQA dataset, which is believed to be more challenging and can demonstrate each model's reasoning ability.
In addition, we test our model's performance on different qualities of scene graph (we manually add different levels of noises) using the validation split to study the impact of semantic graph quality.

\subsection{Implementation details}
\label{section-details}
Visual features are extracted from BUA~\cite{anderson2018bottom}, and the scene graph is built in a similar way as~\cite{DBLP:conf/cvpr/KimKOHZ20}. 
For each bounding box, we select the top-$5$ predictions to constitute its SuperNode.
We experimented with more candidates, but this would greatly increase the computational complexity and the gain is limited and can be even worse.
To better answer direction/location questions, (\ie, left, right, top, bottom), we add two extra nodes that represent the bounding box coordinates (top-left corner $(x_l, y_t)$ and bottom-right corner $(x_l+w, y_t+h)$) for each object in the semantic graph. 
Besides, to reduce the computational load, we merge identical relation and attribute nodes in the semantic graph, and re-build the adjacency matrix accordingly. 
But we do not combine object category nodes. Different same-category nodes match with different identities, which may be critical for question answering, such as counting the number of a particular object. 
The [SEP] token is inserted between the feature sequences from visual ($\{v_{i}\}$ and $\{s_{i}\}$) and question ($\{q_{i}\}$) modalities.

We will release the code once this work is accepted, and more detailed hyper-parameters will be provided in the code repository.
\begin{table*}[!htbp]
	\begin{tabular}{p{0.115\linewidth}p{0.02\linewidth}|ccccc|p{0.112\linewidth}p{0.02\linewidth}|ccc}
		\toprule
		\multirow{2}{*}{\textbf{Method}} & \multirow{2}{*}{\textbf{PT}} & \multicolumn{5}{c|}{\textbf{VQA-v$2^*$}}                                                                          & \multirow{2}{*}{\textbf{Method}} & \multirow{2}{*}{\textbf{PT}} & \multicolumn{3}{c}{\textbf{GQA}}                                        \\ \cline{3-7} \cline{10-12} 
		&                                                                                & \textbf{Y/N} & \textbf{Num.} & \textbf{Other} & \multicolumn{1}{c|}{\textbf{Test-dev}} & \textbf{Test-std} &                                  &                                                                                & \textbf{Open} & \multicolumn{1}{c|}{\textbf{Binary}} & \textbf{Overall} \\ \hline
		BLOCK\footnotesize{\cite{ben2019block}}                            & \xmark                                                                         & 83.6            & 47.33         & 58.51          & \multicolumn{1}{c|}{67.58}             & 67.92             & BUA\footnotesize{\cite{anderson2018bottom}}    & \xmark                                                                         & 34.83         & \multicolumn{1}{c|}{66.64}           & 49.74            \\
		Count\footnotesize{\cite{zhang2018learning}}                       & \xmark                                                                         & 83.14           & 51.62         & 58.97          & \multicolumn{1}{c|}{68.09}             & 83.41             & LRTA\footnotesize{\cite{liang2020lrta}}        & \xmark                                                                         & -             & \multicolumn{1}{c|}{-}               & 54.48            \\
		MCAN\footnotesize{\cite{Yu_2019_CVPR}}             & \xmark                                                                         & 86.82           & 53.26         & 60.72          & \multicolumn{1}{c|}{70.63}             & 70.90             & LCGN\footnotesize{\cite{hu2019language}}                        & \xmark                                                                         & -             & \multicolumn{1}{c|}{-}               & 56.10            \\
		MuRel\footnotesize{\cite{cadene2019murel}}         & \xmark                                                                         & 84.77           & 49.84         & 57.85          & \multicolumn{1}{c|}{68.03}             & 68.41             & LXMERT\footnotesize{\cite{tan2019lxmert}}                       & \cmark                                                                         & -             & \multicolumn{1}{c|}{-}               & 60.33            \\
		Oscar\footnotesize{\cite{li2020oscar}}             & \cmark                                                                         & -               & -             & -              & \multicolumn{1}{c|}{73.14}             & 73.44             & Oscar\footnotesize{\cite{li2020oscar}}                          & \cmark                                                                         & -             & \multicolumn{1}{c|}{-}               & 61.62            \\
		VinVL\footnotesize{\cite{zhang2021vinvl}}          & \cmark                                                                         & -               & -             & -              & \multicolumn{1}{c|}{75.95}             & 76.12             & NSM~\footnotesize{\cite{hudson2019learning}}                     & \xmark                                                                         & 49.25         & \multicolumn{1}{c|}{78.94}           & 63.17            \\
		VL-BERT\footnotesize{\cite{su2019vl}}              & \xmark                                                                         & -               & -             & -              & \multicolumn{1}{c|}{69.58}             & -                 & VinVL\footnotesize{\cite{zhang2021vinvl}}      & \cmark                                                                         & -             & \multicolumn{1}{c|}{-}               & 64.65            \\
		VL-BERT\footnotesize{\cite{su2019vl}}              & \cmark                                                                         & -               & -             & -              & \multicolumn{1}{c|}{71.79}             & 72.22             & Human                            & \xmark                                                                         & 87.40         & \multicolumn{1}{c|}{91.20}           & 89.30            \\ \hline
		SA-VQA                           & \xmark                                                                         & 86.97           & 55.31         & 61.11          & \multicolumn{1}{c|}{71.10}             & 71.38             & SA-VQA                       & \xmark                                                                         & 59.05         & \multicolumn{1}{c|}{77.39}           & 67.65            \\ \bottomrule
		\multicolumn{12}{l}{\footnotesize{*Note that for a fair comparison, the model is not trained on the validation split of the datasets.}}
	\end{tabular}
	\caption{Overall performance comparison on VQA-v2 and GQA datasets.}
	\label{table:overall-acc}
\end{table*}

\subsection{Overall performance}
Experimental results are shown in \tablename~\ref{table:overall-acc}. 
The column ``PT'' indicates whether the method uses external datasets~\cite{NEURIPS2019_98d8a23f,lin2014microsoft,vicente2016large} for pretraining or not.
We compare our method with previous approaches on both the GQA~\cite{DBLP:conf/cvpr/HudsonM19} and VQA-v2~\cite{DBLP:journals/ijcv/AgrawalLAMZPB17} datasets.
It shows that, on GQA dataset, our SA-VQA model gains improvement over the state-of-the-art methods and displays better performance than even the pre-trained models~\cite{tan2019lxmert,li2020oscar,zhang2021vinvl}.
In particular, our approach demonstrates much better accuracy than NSM~\cite{hudson2019learning} on the more challenging \textit{Open} question category.
Meanwhile, on the VQA-v2 dataset, our model outperforms the state-of-the-art methods that do not use extra data for pre-training, and achieve competitive results with the other approaches that employ the pre-training step.

\subsection{Ablation study}
\begin{table}
	\centering
	\begin{tabular}{l|l|l} 
		\toprule
		\textbf{Exp.}  & \textbf{Method}                                                                   & \textbf{Acc.}   \\ 
		\hline
		1            & Two Semantic Trms                                                                 & 56.44           \\ 
		2            & Two Visual Trms                                                                   &
		59.17           \\
		3            & One Trm                                                                           & 61.79           \\
		4            & No Graph Guidance                                                                   & 58.54           \\
		5            & Single Loss on Softmax                                                            & 57.80           \\
		6            & Top-1 Semantic Candidate for Object                                               & 64.44           \\
		7            & Even Softmax (Top-5) Candidates                                                  & 66.67           \\ 
		\hline
		8             & Pretrained SNS (Top-5)(Ours) & 67.65           \\
		\bottomrule
	\end{tabular}
	\caption{Ablation study on GQA dataset. Trm is the abbreviation for Transformer.}
	\label{table:ablation}
\end{table}

We further conduct ablation studies on our SA-VQA model on the GQA test split, and the results are summarized in \tablename~\ref{table:ablation}. 

\subsubsection{Single stream of features from images}
Since SA-VQA uses two transformers to produce the answers, we also employ two transformers in Exp.~1 and Exp.~2 (see \tablename~\ref{table:ablation} to make a fair comparison. 
The two transformers are initialized with different parameters. Again, we utilize late fusion to obtain the final prediction results.
Exp.~1 and Exp.~2 in \tablename~\ref{table:ablation} are their results. 
Compared with the proposed SA-VQA, which uses semantic and visual graphs simultaneously, their performance drops significantly.
This suggests that the semantic or visual branch alone is inadequate for the transformer to infer the correct answers. Especially, as discussed earlier, the semantic graph may be noisy and incomplete, while the visual graph is complete but redundant. Both graphs complement each other and lead to better performance. 
Also, the visual Transformer alone exhibits better performance than the semantic Transformer with the semantic scene graph. 
This may be due to the noises from the scene graph, which can introduce incorrect bounding box labels and attributes. This eventually may mislead the model.
Exp.~5 is designed to test whether fusion is necessary. 
More specifically, we only adopt the last loss term in Eq.~(\ref{eq:loss}) when training the model. The other two loss terms are removed.
The result shows that adding the loss term to each transformer's output is necessary. Otherwise, the model may have much worse performance.
\begin{table}[!t]
	\centering
	\begin{tabular}{l|l|l} 
		\toprule
		\textbf{Exp.} & \textbf{Method}        & \textbf{Accuracy}  \\ 
		\hline
		9  & 60\% SNS Acc  & 67.90     \\
		10  & 70\% SNS Acc  & 69.27     \\
		11  & 80\% SNS Acc  & 76.75     \\
		12  & 90\% SNS Acc  & 79.90     \\
		13  & 100\% SNS Acc & 84.80     \\ 
		\hline
		14  & GT Graph      & 92.64     \\
		\bottomrule
	\end{tabular}
	\caption{Performance on different SNS outputs quality.}
	\label{table:mil}
\end{table}
\begin{figure*}[tb!] 
	\includegraphics[width=2.1\columnwidth]{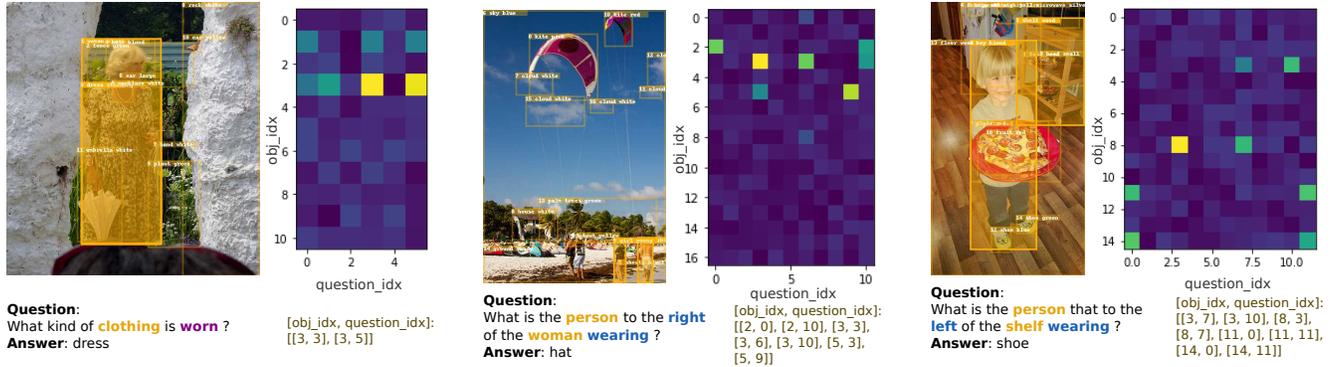}
		\caption{Examples of results from our SA-VQA model. The attention matrix represents the cross-modality attention weights of the last layer. 
			And the colored regions and text represent the highlight attention pairs in the matrix.}
		\label{fig:Transformer_attentions}
	\end{figure*}
	
	\subsubsection{One Transformer for all}
	We also test the merging of all inputs and graphs in one stream (Exp.~3) by expanding the input sequences from three source streams (\ie, semantic graph, visual graph, and question graph) and their corresponding adjacency matrices. 
	The result shows that simultaneously involving multi-modality features in one model is no better than separating them since dividing them into two streams helps the modalities align better. 
	
	\subsubsection{Effectiveness of structured representation $G$}
	Exp.~4 tests the validity of our designed graph by removing all adjacency matrices and guided graphs.
	The result indicates that the Transformer network can achieve decent performance on the GQA dataset, suggesting that it can learn the co-attention between the image and question. 
	However, its performance is still behind the proposed SA-VQA, which further verifies the effectiveness of incorporating structured graph representation into the proposed attention module.
	
	\subsubsection{SuperNodes for objects}
	Exp~.6 and 7 are designed to test the validity of the SuperNode. 
	In Exp.~6, only the top-1 bounding box label is used in the semantic graph, while in Exp.~7 and SA-VQA (Exp.~8), the top-5 bounding-box labels are used.
	Exp.~7 and 8 use different approaches to integrate the top-5 features. 
	Exp.~7 averages the features, which may involve the irrelevant information per bounding box, while Exp.~8 weights the feature by the pre-trained SNS model to eliminate the influence from irrelevant information in top-5 candidates.
	From Exp.~6 and 7, we can also observe that even with extra irrelevant information, better performance is still achieved when using the top-5 semantic candidates than with the top-1 model. 
	This suggests that the graph helps the attention model to discover the related information to the question, which eventually benefits the answer generation.
	Also, from Exp.~7 and 8, we can see that the weighted top-K features contribute more to the overall performance than evenly weighted features.

	{\subsection{Performance on semantic graphs with different levels of accuracy}}
	We pre-train our SNS model on the training split of the GQA dataset, which includes full annotations for the bounding boxes in both training and validation splits. 
	On the validation split, the pre-trained SNS model achieved an accuracy of $47.42\%$. This SNS model is used in our SA-VQA model. 
To test whether a better SNS model can lead to a better overall GQA model, we manually modify the SNS outputs to obtain different levels of prediction accuracy by replacing the ground truth label with a random incorrect label. 
Then, we train different SA-VQA models using the modified SNS outputs.
We randomly divided the validation split into two halves, one for validation and the other for testing.
The results (\tablename~\ref{table:mil}) show that SNS model's performance positively correlates with SA-VQA's performance. 
{This result is expected}, since achieving a better SNS model is equivalent to obtaining better structured representation from the graph. 
Our SA-VQA model can achieve even better results using the ground-truth scene graph, as shown in Exp.~14.

\subsection{Qualitative results}
We also visualize some of the attention maps generated by our SA-VQA model to understand better how the answers are generated. 
Specifically, we visualize the correlations between the image regions and the question words. 
{Some examples are shown in \figurename~\ref{fig:Transformer_attentions}}.
The horizontal axis of the matrix represents the question {word} index, while the vertical {axis} represents the object's index in the image. 
A brighter cell in the matrix indicates a higher correlation between the corresponding object and question word. 
The presented examples show that our model can find the relevant question words in the image, achieving cross-modality alignment and interpretability.
\section{Conclusion and future work}
In this work, we present a novel model named SA-VQA. 
In contrast to the previous work, SA-VQA integrates structured graph information to guide dual Transformers; consequently, it outperforms state-of-the-art methods. 
In addition, we validate our model via our comprehensive ablation studies, showing each component's validity. 
Guidance from the graph significantly improves the overall performance. 
Furthermore, we conduct experiments to test the impact of semantic graphs with different levels of accuracy. Our results demonstrate that a better pre-trained SNS model or a more accurate scene graph improves overall performance. 
The study of integrating structured alignment into VQA is still in its infancy. 
One open issue is pre-training a good semantic graph model to obtain clean and tidy scene information. 
The proposed SA-VQA is a bold attempt towards this issue. However, we expect there will be more research results in the future. 

{\small
\bibliographystyle{ieee_fullname}
\bibliography{PaperForReview}
}

\end{document}